\pdfoutput=1

\documentclass[11pt]{article}

\usepackage[]{ACL2023}

\usepackage{times}
\usepackage{latexsym}
\usepackage{float} 
\usepackage{graphicx}
\usepackage{booktabs,siunitx}

\usepackage[T1]{fontenc}

\usepackage[utf8]{inputenc}

\usepackage{microtype}

\usepackage{inconsolata}

\usepackage{tabularx}

\title{Stanford MLab at SemEval-2023 Task 10: Exploring GloVe- and Transformer-Based Methods for the Explainable Detection of Online Sexism}

\author {
Hee Jung Choi*, Trevor Chow*, Aaron Wan*, Hong Meng Yam*, \\ \textbf{Swetha Yogeswaran*, Beining Zhou*}\\ 
Stanford University \\
\texttt{\{cheejung, tmychow, aaronwan, hongmeng, swethay, cathyzbn\}@stanford.edu}
}

\begin{document}

\maketitle

\def\thefootnote{*}\footnotetext{Authors are listed alphabetically. All authors contributed equally to this work.}\def\thefootnote{\arabic{footnote}}

\begin{abstract}
In this paper, we discuss the methods we applied at SemEval-2023 Task 10: Towards the Explainable Detection of Online Sexism. Given an input text, we perform three classification tasks to predict whether the text is sexist and classify the sexist text into subcategories in order to provide an additional explanation as to why the text is sexist. We explored many different types of models, including GloVe embeddings as the baseline approach, transformer-based deep learning models like BERT, RoBERTa, and DeBERTa, ensemble models, and model blending. We explored various data cleaning and augmentation methods to improve model performance. Pre-training transformer models yielded significant improvements in performance, and ensembles and blending slightly improved robustness in the F1 score.
\end{abstract}

\section{Introduction}
Online sexism has the potential to inflict significant harm on women \cite{ortiz2023}, and it is a serious issue that must be addressed. With the increasing prevalence of social media, it has become easy for groups of people to spread sexist ideas and threaten the safety of others, with online social networks becoming increasingly inundated by sexist comments \cite{Founta2018}. 

There have been numerous previous works on the detection of online sexism as a whole \cite{Schtz2021AutomaticSD, AldanaBobadilla2021ALM, 9281090}, including even applying these models to non-English datasets \cite{Kumar2021SexismDI, Paula2021SexismPI, Jiang2021SWSRAC}. However, almost all of these models do not focus on precisely classifying why a certain text conveys sexist sentiments, and instead provide a binary classification for whether the text is sexist.

However, this is the task which is most likely to make these models useful for content moderation, since they provide both the moderator and the platform's users with a consistent explanation for why something was classified as sexist or not. In many cases, the difficulty of using machine classification when doing content moderation is the perception of it being decided by an arbitrary black box.

While the models may empirically have a high accuracy, due to the high social sensitivity of this task, human intervention is almost always required. With laws such as the General Data Protection Regulation (GDPR) established in Europe, which establishes a "right to explanation"\cite{Hoofnagle2019TheEU}, there is thus a huge need for model explainability on top of existing performance optimization \cite{Mathew2020HateXplainAB}. With more detailed feedback about the categories of sexism, moderators can efficiently mitigate sexist sentiment online in a robust and rules-based manner, and therefore reduce gender-based violence. 

As such, given the increasing importance of explainable detection in machine learning models, we propose and compare several natural language processing methods for doing so. We used GloVe- and transformer-based models, as well as various data cleaning and augmentation techniques, applying them on Reddit and Gab textual data to detect sexist messages and classify them into various categories of sexism. 

\section{Background and Task Setup}
The data for this task was provided by SemEval Task 10 \cite{kirk2022}. This labeled data set consisted of 10,000 entries extracted from Gab along with 10,000 entries from Reddit. The dataset is labeled according to the specifications of the required classifier for subtask A, subtask B, and subtask C. In addition to this labeled dataset, there were two unlabeled data sets which each contain 1 million entries from Gab and Reddit that were provided, which we used to improve our system's performance.

Subtask A requires a binary classifier to categorize posts into being sexist or non-sexist. Subtask B requires a four-class classification system which categorizes a \textit{sexist} post according to one of the following categories: (1) threats, (2) derogation, (3) animosity, and (4) prejudiced discussions. Finally, for subtask C, of the posts which are \textit{sexist}, an 11-class classification system categorizes the posts according to a more specific label of sexism. Subtasks B and C ensure that text which is labeled as sexist is given a specific label for why it is sexist, providing a degree of explainability for those using the model.

\section{System Overview}
\subsection{Data Cleaning}
The training data was taken from Reddit and Gab, and as such, it was essential to clean the data to get consistent formatting. Furthermore, all URL references were removed, hyphens and hashtags were replaced with spaces, all punctuation except apostrophes was removed, and all text was changed to lowercase. Finally, many slang abbreviations were replaced by their expanded forms using the mapping provided by the sms$\_$slang$\_$translator github repository\footnote{https://github.com/rishabhverma17/sms$\_$slang$\_$translator}.
\subsection{Data Augmentation}
\subsubsection{Back Translation}
For Subtask A, since the provided dataset contained far more "Not Sexist" samples than "Sexist" samples, we attempted to use back translation to generate augmented samples of the minority class. Specifically, we translated the minority samples in our training split to Dutch and then back to English, ensuring an even class distribution in our training split. We chose Dutch because it worked empirically \cite{DBLP:journals/corr/abs-2106-04681} and since Dutch is one of the most similar languages to English, owing to their joint lineage in the West Germanic family. Results with back translation are specifically labeled in the results section.

\subsubsection{Easy Data Augmentation}
Since back translation did not improve our results in Subtask A, we attempted a different data augmentation approach for Subtask B: Easy Data Augmentation (EDA). We specifically followed a procedure similar to \cite{kalra2021}. We used three operations–– synonym replacement, random insertion, and random swap with a rate of 0.05––to generate augmented samples of the three minority classes in Subtask B (namely "threats, plans to harm and incitement", "animosity", and "prejudiced discussions") in our training split. Much as with back translation, we generated enough samples for each minority class until the number of samples for each class was equal. Results with EDA are specifically labeled in the results section.

\subsection{GloVe-Based Model}
For our baseline model for Subtask A, we developed a GloVe-based logistic regression model. Although this is not a state-of-the-art model, it provided a useful benchmark for the performance of non-deep learning approaches. We used 50-dimensional GloVe vectors pre-trained on 2 billion tweets from \cite{pennington-etal-2014-glove} to transform each word in the input text into its vector representation. For each sample's input text, we averaged the word vectors across the text to create a 50-dimensional input that we fit with a logistic regression model.

\subsection{Transformer-based Models}
\subsubsection{BERT}
BERT (Bidirectional Encoder Representations for Transformers) is a large-language model that has achieved impressive results in NLP experiments \cite{devlin2018}. It uses a multi-layer, transformer-based encoder architecture and bidirectional self-attention to learn context from both preceding and following sentences. BERT was trained on a language modeling task as well as a next sentence prediction task.

We fine-tuned the BERT model to apply it to our specific Subtasks. We used the existing pre-trained bert-base-uncased tokenizer to preprocess the text for input the BERT model. We added one linear layer with ReLU activation as well as a second linear layer with the Sigmoid (Subtask A) or Softmax (Subtask B and C) activation function to generate the model output. The ultimate prediction was determined either by threshold (for Subtask A) or the argmax of the output vector (for Subtask B and C). For Subtask A, we found that the optimal threshold for a positive prediction was 0.35, as seen in Figure 1. If the model's output was 0.35 or greater, the text would be classified as sexist. We added dropout with a rate of 0.5 to help reduce overfitting.

\begin{figure}[!h]
    \centering
    \includegraphics[scale=0.47]{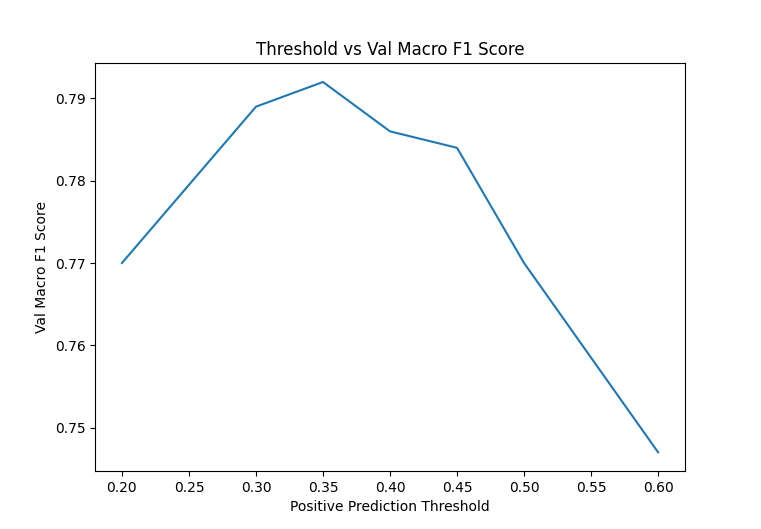}
    \caption{Positive Prediction Thresholds for BERT Classifier}
    \label{fig:my_label}
\end{figure}

\subsubsection{RoBERTa}
RoBERTa (Robustly optimized BERT approach), improves on BERT by using a larger-scale model trained on an even larger and cleaner corpus of text using a longer training schedule, larger batch sizes, and a more advanced masking strategy, resulting in improved performance on a wide range of natural language processing tasks \cite{liu2019}.

We fine-tuned two separate RoBERTa models using the same architecture as our BERT model. We chose to use RoBERTa because it is trained more robustly using dynamic masking as compared to static masking in BERT, and was proven empirically to yield better results. This is because the generation of masks during training means that for each input, the number of different possible masks is much larger than with BERT.

The two separate RoBERTa models included: one existing RoBERTa model that was fine-tuned to classify sexist tweets (ft-RoBERTa)\footnote{https://huggingface.co/annahaz/xlm-roberta-base-misogyny-sexism-tweets}, and one RoBERTa model that was pre-trained on the provided unlabeled data (pt-RoBERTa)\footnote{https://huggingface.co/HPL/roberta-large-unlabeled-labeled-gab-reddit-task-semeval2023-t10-270000sample}.

For Subtask A, we fine-tuned the ft-RoBERTa model. For Subtask B, we fine-tuned both the ft-RoBERTa and the pt-RoBERTa models. For Subtask C, we fine-tuned the pt-RoBERTa model. For all three Subtasks, we used the same architecture as the BERT model.

\subsubsection{HateBERT}

HateBERT is a model that improves upon BERT for the task of abusive language detection \cite{caselli-etal-2021-hatebert}. More specifically, HateBERT uses training data with abusive language to better focus BERT's attention on relevant linguistic cues for this task. It employs a refined pre-training procedure using a large dataset of comments from banned subreddits containing offensive, abusive and hateful language to capture the nuances of this type of speech. Overall, these refinements on domain-specific data have allowed HateBERT to empirically achieve superior performance on various English hate speech detection datasets. We used this only for Subtask B.

\subsubsection{DeBERTa}
DeBERTa (Decoding-enhanced BERT with disentangled attention) is a model that improves upon both BERT and RoBERTa \cite{he2020}. More specifically, DeBERTa uses disentangled attention to better focus on relevant linguistic information, employs a more advanced decoding strategy that captures long-range dependencies, shares parameters between layers to improve efficiency and performance, and uses improved pre-training strategies to capture complex linguistic relationships. Overall, these improvements have led the DeBERTa model to empirically achieve better performance on various downstream NLP tasks.

For Subtasks B and C, we fine-tuned a DeBERTa model for the respective downstream tasks, using the same architecture as the BERT model. 

\subsubsection{Sentence-BERT}
Sentence-BERT (SBERT) is a modification of BERT, with the key innovation being that SBERT is fine-tuned to encode sentences into fixed-length vectors that capture the semantic meaning of the sentence \cite{Reimers2019}. Since SBERT focuses on generating high-quality sentence embeddings while BERT generates embeddings on the word level, we experimented with the SBERT-based model in hopes of addressing the overfitting we saw when applying other BERT-based models. For Subtask B, we fine-tuned the SBERT model using the same architecture we used for the other BERT-based models.

\subsubsection{Ensembles}
In hopes of improving model performance, we experimented with ensemble models that involved concatenating embeddings from two different transformers into a single feature vector and then applying the linear and dropout layers on top of that feature vector to generate the ultimate prediction. We specifically experimented with two ensemble models: concatenating pt-RoBERTa embeddings with DeBERTa embeddings (referred to as Ensemble 1) and concatenating pt-RoBERTa embeddings with SBERT embeddings (referred to as Ensemble 2). Through concatenating the embeddings, we hoped to capture different aspects of the input text that may be better represented by one model over the other and develop a more comprehensive representation of the input text. 

For Subtask B, we applied both Ensemble 1 and Ensemble 2, and for Subtask C, we applied Ensemble 1. For each ensemble model, both transformer embeddings were simultaneously fine-tuned on the corresponding downstream task during model training.

\subsubsection{Model Blending}
To further improve model performance, we experimented with a blending strategy that involved taking the weighted average between the predictions of two different models, giving slightly greater weight to the better-performing model. For Subtask A, we took the weighted average of the ft-RoBERTa and BERT models, multiplying ft-RoBERTa's prediction by 0.6 and BERT's prediction by 0.4 and summing the results for the ultimate prediction. For Subtasks B and C, we took the weighted average of Ensemble 1 and the pt-RoBERTa model. Similar to Subtask A's blended model, we multiplied Ensemble 1's predictions by 0.6 and pt-RoBERTa's predictions by 0.4 and summed them to generate the ultimate prediction.

\section{Experimental Setup}
To train and evaluate our models, we used an 80-20 split on the provided training dataset to create training and validation datasets for our models. We used a fixed train-val split so we could directly compare the performance of our models. During training, we monitored the Macro F1 score on the validation set, and our program saved the model with the best score. We trained our models with a maximum of 200 epochs, and we used early-stopping to stop training if the training loss did not decrease over 30 epochs. We trained all of our transformer-based models using the Adam optimizer and cross-entropy loss.

For our final submission, we trained our model on all the training data provided.

\section{Results}
\subsection{Summary}
As seen in Table 1, the best-performing model for Subtask A was the weighted-average blending strategy, though the difference in F1 scores between the three best-performing models (BERT, ft-RoBERTa, and Weighted Average) was small. ft-RoBERTa had a slightly better score than the plain BERT model. GloVe Vector-based model yielded the worse performance. The data augmentation we performed, such as back-translation, did not lead to any improvements.

For Subtask B, we see in Table 2 that pt-RoBERTa clearly sets itself from the pack out of all transformer-based models, achieving an F1 of 0.62. This is very close to the best results achieved by the models involving multiple transformers, as Ensemble 1 and the Weighted Average model achieved F1s of 0.622 and 0.624, respectively. Like with back translation in Subtask A, the EDA strategy we used for Subtask B failed to lead to any improvements. HateBERT, which is pre-trained on abusive text, beat out most of the benchmark BERT-based models except pre-trained RoBERTa, suggesting that the features it learns from abusive text do not perfectly translate across to sexism. 

To prove the importance of data cleaning, we ran an experiment with our pt-RoBERTa model on the uncleaned input text, as seen in Table 2. The resulting F1 score of 0.571 was a significant decrease from the score on the cleaned input text.

Our results in Table 3 show that Ensemble 1 outperforms pt-RoBERTa and the Weighted Average model in Subtask C, which was a surprising difference given the results from Subtask B.

The performance of our final models used for submission on the Dev and Test sets can be seen in Table 4 and Table 5, respectively. Our models' performances on the Dev set were relatively consistent with our Val set results, but the performance on the Test set represented a noticeable decline, especially for Subtasks B and C.

\subsection{Discussion}
We see from Subtask A that ft-RoBERTa yielded slightly better results than the plain BERT model. However, since we did not test a plain RoBERTa model and because the difference in performance is very small, it is difficult to tell if this improvement was primarily due to the improvements of RoBERTa over BERT or the transfer-learning from fine-tuning on the Twitter task.

However, when it comes to pre-training on domain-specific data, we can clearly see that this is vital to improving results. In Subtask B, we see that pt-RoBERTa outperformed all single-transformer models, including DeBERTa. This shows that the success of pt-RoBERTa model can be primarily attributed to the more robust embeddings created after pre-training the RoBERTa embeddings on the unlabeled dataset.

For Subtasks B and C, creating ensemble-type models by concatenating embeddings from different transformer models produced slight improvements. Ensemble 2 was middle-of-the-pack in Subtask B since it involved concatenating embeddings from a poorer-performing transformers mode (SBERT). For Ensemble 1, while the improvement was only slight for Subtask B, there is a clear difference in performance between the pt-RoBERTa model's F1 score and Ensemble 1's F1 score for Subtask C. This demonstrates how concatenating embeddings from different transformer models can be an effective strategy for creating more robust representations of the input text.

For Subtasks A and B, blending the predictions of the best-performing models led to slight improvements in performance. However, the improvements were small, and for Subtask C, blending the models did not improve results. This indicates that model blending may not be the most optimal approach to improving model performance.

\begin{table}[!h]
\captionof{table}{Val Macro F1 scores of Subtask A Models}
\begin{tabular}{@{}lr@{}}
\toprule
\multicolumn{1}{c}{\textbf{Model}}             & \multicolumn{1}{c}{\textbf{Val F1}} \\ \midrule
GloVe Vectors + Logistic Regression            & 0.623                               \\
BERT                                           & 0.792                               \\
ft-RoBERTa                                     & 0.798                               \\
Weighted Average: \\ BERT \& ft-Roberta           & 0.805                               \\
Augmentation: \\ BERT \& Back Translation       & 0.789
\end{tabular}
\end{table}

\begin{table}[!h]
\captionof{table}{Val Macro F1 scores of Subtask B Models}
\begin{tabular}{@{}lr@{}}
\toprule
\multicolumn{1}{c}{\textbf{Model}}                                                                & \multicolumn{1}{c}{\textbf{Val F1}} \\ \midrule
SBERT                                                                             & 0.534                               \\
BERT                                                                                      & 0.521                               \\
DeBERTa                                                                                     & 0.562                               \\
HateBERT & 0.582\\
ft-RoBERTa                                                                                        & 0.525                               \\
pt-RoBERTa                                                                                        & 0.62                                \\
\begin{tabular}[c]{@{}l@{}}Ensemble 2 \end{tabular}      & 0.555                               \\
\begin{tabular}[c]{@{}l@{}}Ensemble 1\end{tabular}               & 0.622                               \\
\begin{tabular}[c]{@{}l@{}}Weighted Average: \\En. 1 $\&$ pt-RoBERTa\end{tabular} & 0.624              \\
Augmented: \\ pt-RoBERTa \& EDA & 0.618 \\
Uncleaned: \\ pt-RoBERTa \& Raw Data & 0.571
               \\ \bottomrule
\end{tabular}
\end{table}

\begin{table}[!h]
\captionof{table}{Val Macro F1 scores of Subtask C Models}
\begin{tabular}{@{}lr@{}}
\toprule
\multicolumn{1}{c}{\textbf{Model}}                                                                 & \multicolumn{1}{c}{\textbf{Val F1}} \\ \midrule
pt-RoBERTa                                                                          & 0.393                               \\
\begin{tabular}[c]{@{}l@{}}Ensemble 1\end{tabular} & 0.416          \\                   
\begin{tabular}[c]{@{}l@{}}Weighted Average: \\En. 1 $\&$ pt-RoBERTa \end{tabular} & 0.405

\end{tabular}

\end{table}

\begin{table}[!h]
\captionof{table}{Dev Set Results}
\begin{tabular}{@{}lr@{}}
\toprule
\multicolumn{1}{c}{\textbf{Model}}                                                                 & \multicolumn{1}{c}{\textbf{Dev F1}} \\ \midrule
\begin{tabular}[c]{@{}l@{}}Weighted Average: \\ BERT \& ft-RoBERTa (A)\end{tabular} & 0.802                             \\
\begin{tabular}[c]{@{}l@{}}Weighted Average: \\ Ensemble 1 $\&$ pt-RoBERTa (B) \end{tabular} & 0.628
\\
\begin{tabular}[c]{@{}l@{}}Ensemble 1 (C) \end{tabular} & 0.382

\end{tabular}

\end{table}

\begin{table}[!h]
\captionof{table}{Test Set Results}
\begin{tabular}{@{}lr@{}}
\toprule
\multicolumn{1}{c}{\textbf{Model}}                                                                 & \multicolumn{1}{c}{\textbf{Test F1}} \\ \midrule  
\begin{tabular}[c]{@{}l@{}}Weighted Average: \\ BERT \& ft-RoBERTa (A)\end{tabular}  & 0.798
\\
\begin{tabular}[c]{@{}l@{}}Weighted Average: \\ Ensemble 1 $\&$ pt-RoBERTa (B) \end{tabular} & 0.573
\\
\begin{tabular}[c]{@{}l@{}}Ensemble 1 (C) \end{tabular} & 0.354

\end{tabular}

\end{table}

\pagebreak
\section{Conclusion}
The three subtasks and classification models for online sexism were able to provide explanations for the model predictions by showing intermediate results of the classification. Although this does not provide a look inside a black box, it is nonetheless a useful explanation for the end user of the model, explaining the specific reason why a text might be sexist.

From our experiments, we saw that transformer-based models like BERT and RoBERTa worked the best to classify sexism in texts, as seen in Subtask A. Data cleaning was essential in improving our results. Furthermore, pre-training transformers models like RoBERTa on domain-specific text substantially improved the performance, rivaling the multi-transformer models in Subtask B. 

The models involving concatenating transformer embeddings produced slightly (Subtask B) to significantly better (Subtask C) results, illustrating how combining information from different transformer models produces better representations. Blending model outputs led to slight performance improvements, but these improvements were small in comparison to the improvements seen from pre-training and concatenation.

\section{Limitations and Future Work}

In this paper, we chose the pre-trained RoBERTa model as a consistent benchmark across all three tasks. Beyond RoBERTa, we used differing models for different tasks by iterating on their performances in previous stages, introducing new models (e.g. SBERT, DeBERTa, HateBERT etc.) for Task B due to the poor performance of some models in Task A and picking the better-performing models in Task B for Task C.

Given the significant improvements shown in the pre-trained RoBERTa model on the unlabeled data, our system could be further improved if we pre-trained more robust models, such as DeBERTa, on the unlabeled data. We would like to explore this direction in the future if we had more computational resources.

Ensembles and model blending led to slight performance improvements. However, there are still many combinations and methods of transformer ensembles and model blending we were unable to experiment with due to time constraints. To build a more robust model, we would systematically experiment with other techniques and combinations of ensembles and model blending.

The data augmentation approaches we attempted failed to lead to better results. This could be due to the fact that the augmented data is not in the distribution of the dataset, or it could show that our model is slightly overfitted. We should like to explore other augmentation strategies since dealing with minority classes would be key in further improving the macro F1 score of our system. To navigate class imbalance, we would like to experiment with more generative models in addition to data augmentation and weighting methods.

Nonetheless, we believe these results show the potential of using pre-trained transformer models coupled with concatenating embeddings in explainable textual detection.

\section*{Acknowledgements}
This research effort would not have been possible without the support of Stanford ACMLab. We would also like to thank Hannah Rose Kirk, Wenjie Yin, Paul Röttger, and Dr. Bertie Vidgen for organizing SemEval 2023 Task 10: Towards the Explainable Detection of Online Sexism. Furthermore, we would like to thank the reviewers for their insightful comments which strengthened the quality of our paper. Finally, we would like to acknowledge Google Colaboratory for its free computing services.

\bibliography{anthology,custom}

\begin{thebibliography}{19}
\expandafter\ifx\csname natexlab\endcsname\relax\def\natexlab#1{#1}\fi

\bibitem[{Aldana-Bobadilla et~al.(2021)Aldana-Bobadilla, Molina-Villegas,
  Montelongo-Padilla, Lopez-Arevalo, and Sordia}]{AldanaBobadilla2021ALM}
Edwin Aldana-Bobadilla, Alejandro Molina-Villegas, Yuridia Montelongo-Padilla,
  I.~Lopez-Arevalo, and Oscar~S. Sordia. 2021.
\newblock A language model for misogyny detection in latin american spanish
  driven by multisource feature extraction and transformers.
\newblock \emph{Applied Sciences}.

\bibitem[{Beddiar et~al.(2021)Beddiar, Jahan, and
  Oussalah}]{DBLP:journals/corr/abs-2106-04681}
Djamila~Romaissa Beddiar, Md~Saroar Jahan, and Mourad Oussalah. 2021.
\newblock \href {http://arxiv.org/abs/2106.04681} {Data expansion using back
  translation and paraphrasing for hate speech detection}.
\newblock \emph{CoRR}, abs/2106.04681.

\bibitem[{Caselli et~al.(2021)Caselli, Basile, Mitrovi{\'c}, and
  Granitzer}]{caselli-etal-2021-hatebert}
Tommaso Caselli, Valerio Basile, Jelena Mitrovi{\'c}, and Michael Granitzer.
  2021.
\newblock \href {https://doi.org/10.18653/v1/2021.woah-1.3} {{H}ate{BERT}:
  Retraining {BERT} for abusive language detection in {E}nglish}.
\newblock In \emph{Proceedings of the 5th Workshop on Online Abuse and Harms
  (WOAH 2021)}, pages 17--25, Online. Association for Computational
  Linguistics.

\bibitem[{de~Paula et~al.(2021)de~Paula, da~Silva, and
  Schlicht}]{Paula2021SexismPI}
Angel Felipe~Magnoss{\~a}o de~Paula, Roberto~Fray da~Silva, and I.~Baris
  Schlicht. 2021.
\newblock Sexism prediction in spanish and english tweets using monolingual and
  multilingual bert and ensemble models.
\newblock \emph{ArXiv}, abs/2111.04551.

\bibitem[{Devlin et~al.(2018)Devlin, Chang, Lee, and Toutanova}]{devlin2018}
Jacob Devlin, Ming{-}Wei Chang, Kenton Lee, and Kristina Toutanova. 2018.
\newblock {BERT:} pre-training of deep bidirectional transformers for language
  understanding.
\newblock \emph{CoRR}, abs/1810.04805.

\bibitem[{Founta et~al.(2018)Founta, Djouvas, Chatzakou, Leontiadis, Blackburn,
  Stringhini, Vakali, Sirivianos, and Kourtellis}]{Founta2018}
Antigoni-Maria Founta, Constantinos Djouvas, Despoina Chatzakou, Ilias
  Leontiadis, Jeremy Blackburn, Gianluca Stringhini, Athena Vakali, Michael
  Sirivianos, and Nicolas Kourtellis. 2018.
\newblock Large scale crowdsourcing and characterization of twitter abusive
  behavior.
\newblock \emph{Proceedings of the international AAAI conference on web and
  social media (ICWSM)}, abs/1802.00393.

\bibitem[{He et~al.(2020)He, Liu, Gao, and Chen}]{he2020}
Pengcheng He, Xiaodong Liu, Jianfeng Gao, and Weizhu Chen. 2020.
\newblock Deberta: Decoding-enhanced {BERT} with disentangled attention.
\newblock \emph{CoRR}, abs/2006.03654.

\bibitem[{Hoofnagle et~al.(2019)Hoofnagle, van~der Sloot, and
  Borgesius}]{Hoofnagle2019TheEU}
Chris~Jay Hoofnagle, Bart van~der Sloot, and Frederik J.~Zuiderveen Borgesius.
  2019.
\newblock The european union general data protection regulation: what it is and
  what it means*.
\newblock \emph{Information \& Communications Technology Law}, 28:65 -- 98.

\bibitem[{Jiang et~al.(2021)Jiang, Yang, Liu, and Zubiaga}]{Jiang2021SWSRAC}
Aiqi Jiang, Xiaohan Yang, Yang Liu, and Arkaitz Zubiaga. 2021.
\newblock Swsr: A chinese dataset and lexicon for online sexism detection.
\newblock \emph{Online Soc. Networks Media}, 27:100182.

\bibitem[{Kalra and Zubiaga(2021)}]{kalra2021}
Amikul Kalra and Arkaitz Zubiaga. 2021.
\newblock Sexism identification in tweets and gabs using deep neural networks.
\newblock \emph{CoRR}, abs/2111.03612.

\bibitem[{Kirk et~al.(2022)Kirk, Yin, Röttger, and Vidgen}]{kirk2022}
Hannah~Rose Kirk, Wenjie Yin, Paul Röttger, and Bertie Vidgen. 2022.
\newblock Semeval 2023 task 10: Towards the explainable detection of online
  sexism (edos).

\bibitem[{Kumar et~al.(2021)Kumar, Pal, and Pamula}]{Kumar2021SexismDI}
Ritesh Kumar, Soumya Pal, and Rajendra Pamula. 2021.
\newblock Sexism detection in english and spanish tweets.
\newblock In \emph{IberLEF@SEPLN}.

\bibitem[{Liu et~al.(2019)Liu, Ott, Goyal, Du, Joshi, Chen, Levy, Lewis,
  Zettlemoyer, and Stoyanov}]{liu2019}
Yinhan Liu, Myle Ott, Naman Goyal, Jingfei Du, Mandar Joshi, Danqi Chen, Omer
  Levy, Mike Lewis, Luke Zettlemoyer, and Veselin Stoyanov. 2019.
\newblock Roberta: {A} robustly optimized {BERT} pretraining approach.
\newblock \emph{CoRR}, abs/1907.11692.

\bibitem[{Mathew et~al.(2020)Mathew, Saha, Yimam, Biemann, Goyal, and
  Mukherjee}]{Mathew2020HateXplainAB}
Binny Mathew, Punyajoy Saha, Seid~Muhie Yimam, Chris Biemann, Pawan Goyal, and
  Animesh Mukherjee. 2020.
\newblock Hatexplain: A benchmark dataset for explainable hate speech
  detection.
\newblock In \emph{AAAI Conference on Artificial Intelligence}.

\bibitem[{Ortiz(2023)}]{ortiz2023}
Stephanie Ortiz. 2023.
\newblock “if something ever happened, i’d have no one to tell:” how
  online sexism perpetuates young women’s silence.
\newblock \emph{Feminist Media Studies}, 1-16.

\bibitem[{Pennington et~al.(2014)Pennington, Socher, and
  Manning}]{pennington-etal-2014-glove}
Jeffrey Pennington, Richard Socher, and Christopher Manning. 2014.
\newblock \href {https://doi.org/10.3115/v1/D14-1162} {{G}lo{V}e: Global
  vectors for word representation}.
\newblock In \emph{Proceedings of the 2014 Conference on Empirical Methods in
  Natural Language Processing ({EMNLP})}, pages 1532--1543, Doha, Qatar.
  Association for Computational Linguistics.

\bibitem[{Reimers and Gurevych(2019)}]{Reimers2019}
Nils Reimers and Iryna Gurevych. 2019.
\newblock Sentence-bert: Sentence embeddings using siamese bert-networks.
\newblock \emph{CoRR}, abs/1908.10084.

\bibitem[{Rodríguez-Sánchez et~al.(2020)Rodríguez-Sánchez, Carrillo-de
  Albornoz, and Plaza}]{9281090}
Francisco Rodríguez-Sánchez, Jorge Carrillo-de Albornoz, and Laura Plaza.
  2020.
\newblock \href {https://doi.org/10.1109/ACCESS.2020.3042604} {Automatic
  classification of sexism in social networks: An empirical study on twitter
  data}.
\newblock \emph{IEEE Access}, 8:219563--219576.

\bibitem[{Sch{\"u}tz et~al.(2021)Sch{\"u}tz, Boeck, Liakhovets, Slijepcevic,
  Kirchknopf, Hecht, Bogensperger, Schlarb, Schindler, and
  Zeppelzauer}]{Schtz2021AutomaticSD}
Mina Sch{\"u}tz, Jaqueline Boeck, Daria Liakhovets, Djordje Slijepcevic, Armin
  Kirchknopf, Manuel Hecht, Johannes Bogensperger, Sven Schlarb, Alexander
  Schindler, and Matthias Zeppelzauer. 2021.
\newblock Automatic sexism detection with multilingual transformer models.
\newblock \emph{ArXiv}, abs/2106.04908.

\end{thebibliography}
\bibliographystyle{acl_natbib}

\end{document}